\documentclass{article}
\usepackage{ijcai22}

\usepackage{times}
\usepackage{soul}
\usepackage{url}
\usepackage[hidelinks]{hyperref}
\usepackage[utf8]{inputenc}
\usepackage[small]{caption}
\usepackage{graphicx}
\usepackage{amsmath}
\usepackage{amsthm}
\usepackage{booktabs}
\usepackage{algorithm}
\usepackage{algorithmic}

\usepackage{tikz}
\usepackage{comment}
\usepackage{amsmath,amssymb} % define this before the line numbering.
\usepackage{color}

\usepackage{graphicx}
\usepackage{amsmath}
\usepackage{amssymb}
\usepackage{booktabs}
\usepackage{amsmath}
\usepackage{tabularx}

\title{Positional Label for Self-Supervised Vision Transformer}

\author{
Zhemin Zhang$^1$
\and
Xun Gong$^1$
\affiliations
$^1$Southwest Jiaotong University, China
\emails
zheminzhang@my.swjtu.edu.cn
}

\begin{document}

\maketitle

\begin{abstract}
Positional encoding is important for vision transformer (ViT) to capture the spatial structure of the input image. General effectiveness has been proven in ViT. In our work we propose to train ViT to recognize the positional label of patches of the input image, this apparently simple task actually yields a meaningful self-supervisory task. Based on previous work on ViT positional encoding, we propose two positional labels dedicated to 2D images including absolute position and relative position. Our positional labels can be easily plugged into various current ViT variants. It can work in two ways: \textbf{(a)} As an auxiliary training target for vanilla ViT (e.g., ViT-B \cite{DBLP:journals/corr/abs-2010-11929} and Swin-B \cite{Liu_2021_ICCV}) for better performance. \textbf{(b)} Combine the self-supervised ViT (e.g., MAE \cite{DBLP:journals/corr/abs-2111-06377}) to provide a more powerful self-supervised signal for semantic feature learning. Experiments demonstrate that with the proposed self-supervised methods, ViT-B and Swin-B gain improvements of 1.20\% (top-1 Acc) and 0.74\% (top-1 Acc) on ImageNet, respectively, and 6.15\% and 1.14\% improvement on Mini-ImageNet.
\end{abstract}

\section{Introduction}

Vision transformers (ViT) \cite{DBLP:journals/corr/abs-2010-11929,Zhao2020CVPR} have recently emerged as an alternative to convolutional neural networks (CNNs) in computer vision. The design of ViT is inspired by the transformer of natural language processing (NLP), split an image into patches and provide the sequence of linear embeddings of these patches as an input to a Transformer \cite{pmlr-v139-touvron21a}. Image patches are treated the same way as tokens (words) in NLP. The core of transformer is self-attention \cite{NIPS2017_3f5ee243}, which is able to model long-range dependencies in the data. However, self-attention is fully symmetric, it cannot capture the ordering of input tokens, which is undesirable for modeling structured data. Therefore, incorporating positional encoding is especially important for ViT.

\begin{figure}[t]
  \centering
   \includegraphics[width=1.0\linewidth]{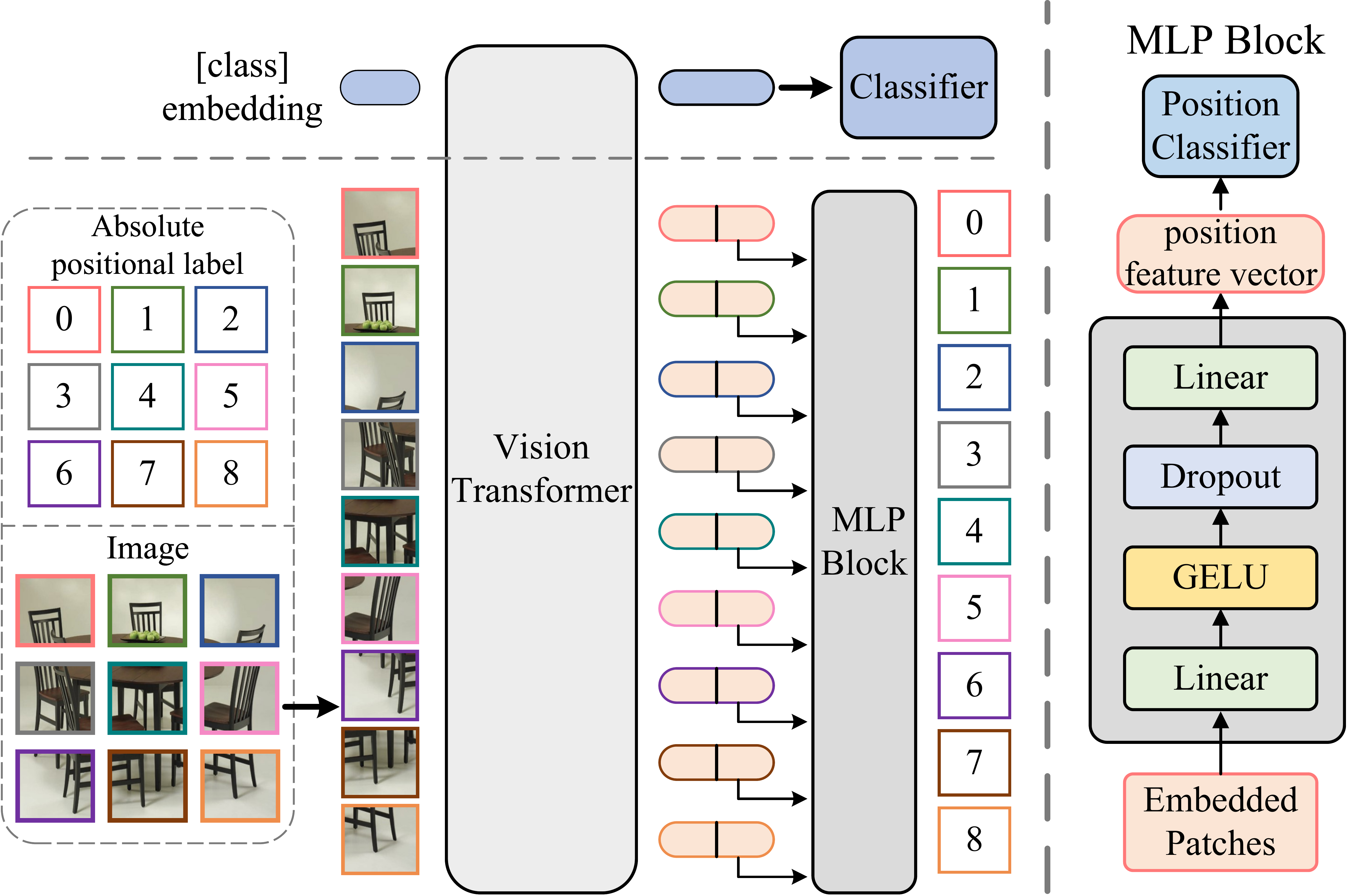}

   \caption{Our absolute positional label combined with ViT-B. We split an image into ﬁxed-size patches, each corresponding to an absolute position, and feed the sequence of image patches to a standard ViT encoder. The full set of encoded patches output (intercept half) by the ViT is processed by a lightweight MLP block that outputs the positional feature vector corresponding to each image patch, which is finally fed into a positional classifier for training, the same as for vanilla classification. After training, the MLP block is discarded.}
   \label{absolute position-flabel}
\end{figure}

There are mainly two classes of methods to add positional information for ViT. One is absolute and the other is relative. Absolute positional encoding \cite{pmlr-v70-gehring17a,Li_2018_CVPR}: each absolute position of the input token sequence from 1 to maximum sequence length has a separate encoding vector. The encoding vector is then combined with the input tokens, usually element-wise add, to incorporate positional information into the ViT's input tokens. On the other hand, relative positional encoding \cite{DBLP:journals/corr/abs-1901-02860,DBLP:journals/corr/abs-1803-02155} calculates the relative position between tokens based on the absolute position of each input token, and establishes a look-up table. Each relative position corresponds to a learnable parameter of the table to learn the pairwise relations of tokens. The parameters in the table interact with the self-attention weights. Relative positional encoding has been veriﬁed to be effective in ViT \cite{Wu_2021_ICCV}. Different from the previous explicit incorporation of positional information, we use the positional information as a supervision signal for ViT self-supervised training, making each encoded patch implicitly contains its positional information, as shown in Figure \ref{absolute position-flabel}.

In this paper, we seek to expand the applicability of positional information to serve as a supervised signal for self-supervised training of ViT. Based on the positional information, we propose a new ViT self-supervised loss function, namely position loss, which allows each token to implicitly contain its positional information. Specifically, we split an image into fixed-size patches and feed the sequence of image patches to a standard ViT. The full set of encoded patches output by the ViT is processed by a lightweight MLP block that outputs the positional label corresponding to each image patch. Based on previous research on ViT positional encoding, we propose two positional labels for ViT: absolute and relative positional labels. Section 3 illustrates how to apply these two positional labels to ViT and their differences.

Our position loss can work in two ways: \textbf{(a)} Combined with vanilla ViT (e.g., ViT-B \cite{DBLP:journals/corr/abs-2010-11929} and swin transformer \cite{Liu_2021_ICCV}), as an auxiliary training task for the model. The ViT is trained under the joint supervision of the softmax loss and position loss, with a hyperparameter to balance the two supervision signals. We find that with joint supervision, the classification performance of ViT is significantly improved. \textbf{(b)} Combined with self-supervised ViT, e.g., MAE \cite{DBLP:journals/corr/abs-2111-06377}. MAE masks random patches from the input image and reconstructs the missing patches in the pixel space. Our position loss combined with MAE, allowing the model to not only reconstructs the pixels of the missing patches, but also outputs the positional information of the patches, providing a more powerful self-supervised signal for semantic feature learning. Our contributions can be summarized as:
\begin{itemize}
\item We propose two positional labels, absolute and relative positional labels, for ViT self-supervised learning. We also introduce efficient implementations of these two positional labels that can be easily plugged into self-attention layers.
\item Our proposed positional label can be used as an alternative for positional encoding, which implicitly adds positional information to each image patch through ViT self-supervised training.
\item The proposed positional labels can be combined with vanilla ViT and self-supervised ViT. Experiments show that, without adjusting any hyperparameters and settings, ViT-B and Swin-B obtained improvements of 1.20\% (top-1 Acc) and 0.74\% (top-1 Acc) on ImageNet, respectively. More improvements on the small dataset, ViT-B and Swin-B obtained improvements of 6.15\% (top-1 Acc) and 1.14\% (top-1 Acc) on Mini-ImageNet, respectively.
\end{itemize}

\section{Related Work}

\subsection{Self-Supervised ViT}
In pioneering works \cite{pmlr-v119-chen20s,DBLP:journals/corr/abs-2111-06377}, training self-supervised Transformers for vision problems in general follows the masked auto-encoding paradigm in NLP \cite{DBLP:journals/corr/abs-1810-04805,NEURIPS2019_4496bf24}. iGPT \cite{pmlr-v119-chen20s} masks and reconstructs pixels. MAE \cite{DBLP:journals/corr/abs-2111-06377} masks a high proportion of the input image patches and reconstruct the missing patches. DINO \cite{Caron_2021_ICCV} studies the importance of momentum encoder, multi-crop training \cite{NEURIPS2020_70feb62b}, and the use of small patches with ViTs, to design a simple self-supervised approach that can be interpreted as a form of knowledge distillation with no labels. \cite{Chen_2021_ICCV} focuses on training Transformers in the contrastive/Siamese paradigm, in which the loss is not deﬁned for reconstructing the inputs. DILEMMA \cite{sameni2022dilemma} proposes a pseudo-task to train a ViT to detect which input token has been combined with an incorrect positional embedding to boost both shape and texture discriminability in models trained via self-supervised learning. In this work, we propose a novel self-supervised learning signal for ViT, named positional labels.

\subsection{Positional Encoding}
\noindent \textbf{(a) Absolute Positional Encoding.} Since transformer contains no recurrence and no convolution, in order for the model to make use of the order of the sequence, we need to inject some information about the position of the tokens. The original self-attention considers the absolute position, and add the absolute positional encodings to the input token embedding by element-wise addition \cite{NIPS2017_3f5ee243}. There are several choices of absolute positional encodings, such as Sinusoidal position \cite{NIPS2017_3f5ee243} fixed encoding using sine and cosine functions sampling at different frequencies and \cite{pmlr-v70-gehring17a} the learnable encoding via training parameters.

\noindent \textbf{(b) Relative Positional Encoding.} Relative positional encoding is proposed ﬁrstly by \cite{DBLP:journals/corr/abs-1803-02155}, extending the self-attention mechanism to efficiently consider representations of the relative positions, where relative positional encodings are added into the self-attention weight calculation. \cite{DBLP:journals/corr/abs-1901-02860} proposed a novel positional encoding scheme with the prior of the sinusoid matrix and more learnable parameters, which not only enables capturing longer-term dependency, but also resolves the context fragmentation problem. \cite{Wu_2021_ICCV} proposes new relative positional encoding methods dedicated to 2D images, called image RPE (iRPE). iRPE considers directional relative distance modeling as well as the interactions between queries and relative positional embeddings in self-attention mechanism. \cite{NEURIPS2019_3416a75f} proposed 2D relative positional encoding that computes and concatenates separate encodings of each dimension. \cite{Srinivas_2021_CVPR} proposes to incorporate the positional information in the convolutional feature map into ViT by concatenating the convolutional feature map with a set of feature maps produced via the self-attentive mechanism. 

Unlike the previous explicit combination of positional information, we use the positional information as a supervised signal for ViT self-supervised training, such that each encoded patch implicitly contains its positional information. Section 4.2 discusses the relationship between our positional label and conventional positional encoding and demonstrates that they can be used jointly.

\begin{figure*}[t]
\centering
\includegraphics[width=0.9\linewidth]{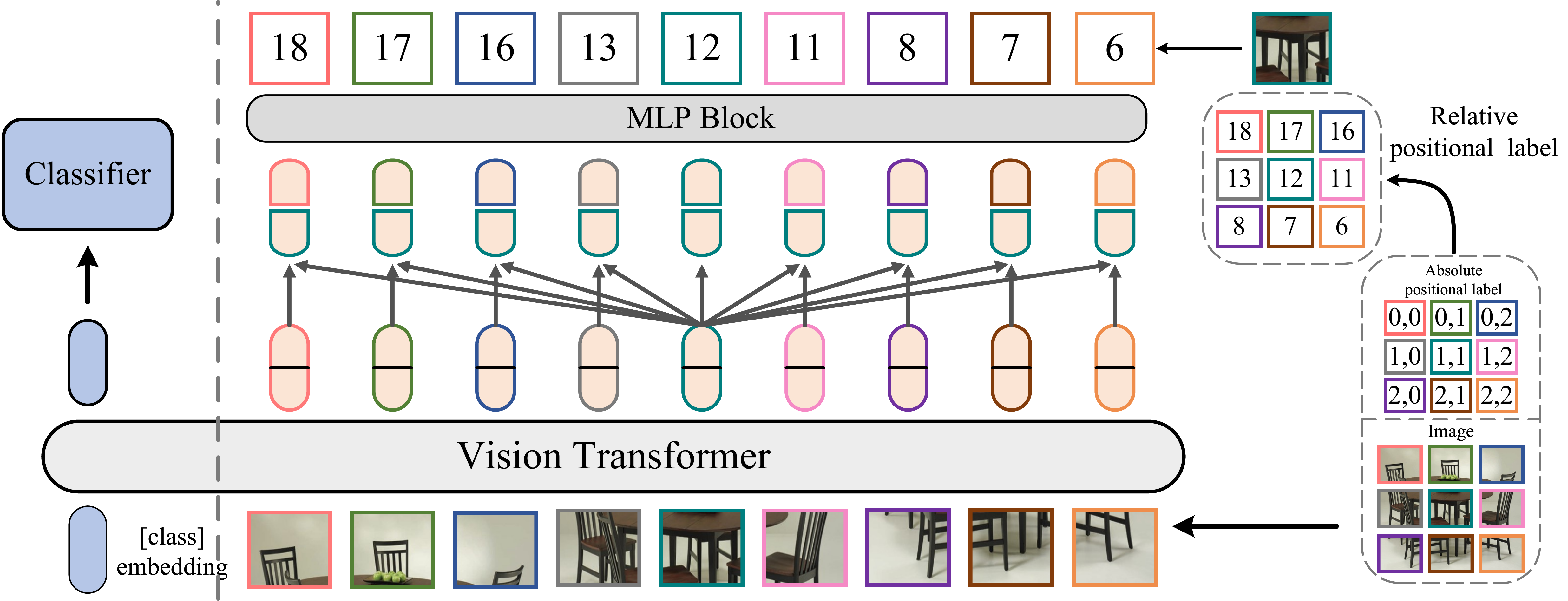} % Reduce the figure size so that it is slightly narrower than the column.
\caption{Our relative positional label combined with ViT-B. For the brevity of the image, here we take the relative positional label of the central image patch of the input image as an example, and the rest of the image patches are calculated in the same way. We combine the center encoded patch with other encoded patches pairwise, input a lightweight MLP block, and output the relative positional labels between the center image patch and other image patches. We use the method in Swin-ViT to convert 2D absolute positional labels to 1D relative positional labels}
\label{relative position-flabel}
\end{figure*}

\begin{figure*}[t]
\centering
\includegraphics[width=0.9\linewidth]{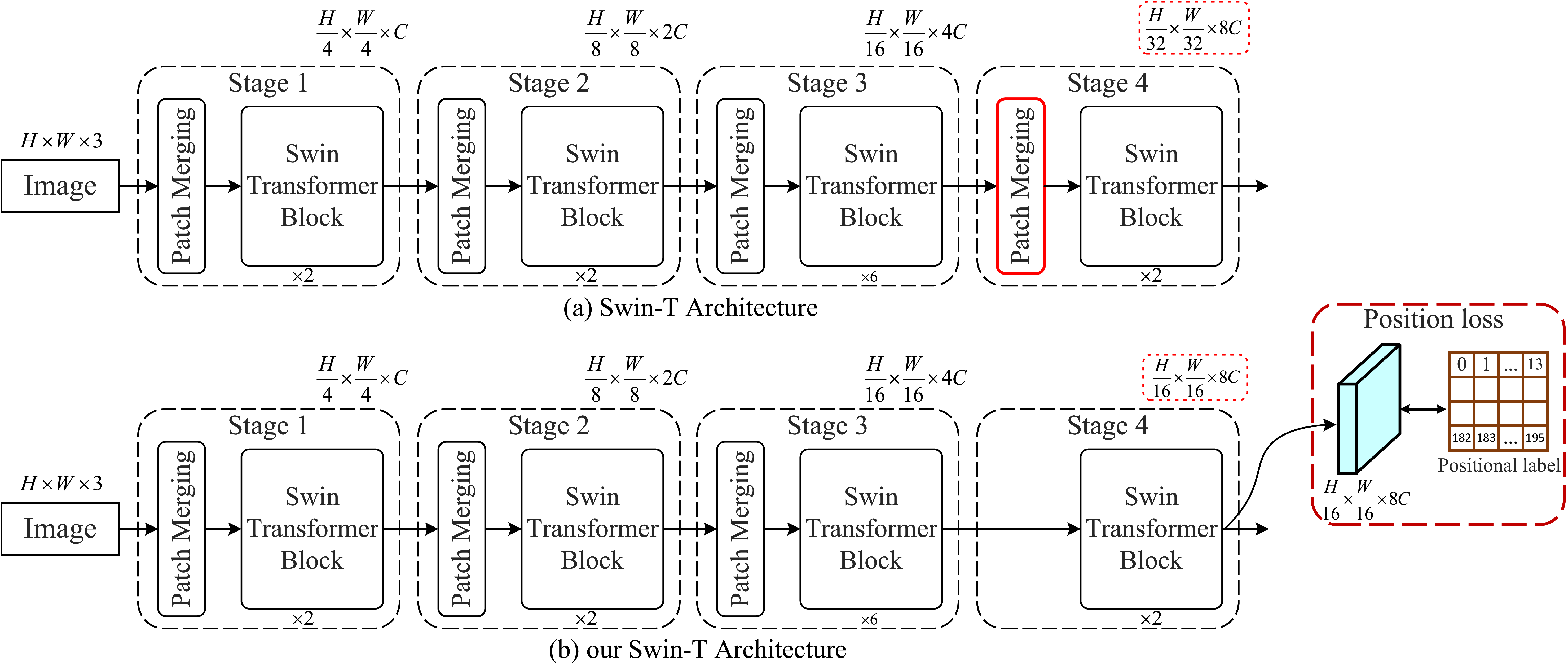} % Reduce the figure size so that it is slightly narrower than the column.
\caption{(a) The architecture of the original Swin Transformer (Swin-T); (b) The architecture of our modified Swin-T. This illustration was inspired by Swin-ViT figure 3(a).}
\label{modified SwinT-flabel}
\end{figure*}

\section{Method}

\subsection{Absolute Positional Label} 

Here we take ViT-B \cite{DBLP:journals/corr/abs-2010-11929} as an example, other ViT variants (e.g., Swin-B \cite{Liu_2021_ICCV}) are calculated in the same way. An overview of the absolute position as an auxiliary training task for classification is depicted in Figure \ref{absolute position-flabel}. 

We split an image $\mathsf{x}\in {{\mathbb{R}}^{H\times W\times C}}$ into fixed-size patches ${{\mathsf{x}}_{m}}\in {{\mathbb{R}}^{N\times ({{m}^{2}}\cdot C)}}$, where $\left( H,W \right)$ is the resolution of the original image, $C$ is the number of channels, $\left( m,m \right)$ is the resolution of each image patch, and $N=HW/{{m}^{2}}$ is the number of generated patches. ${{\mathsf{x}}_{class}}$ is the classification token, whose state at the output of the $\mathsf{ViT}$ serves as the image representation ${{\mathsf{z}}_{class}}$ (Eq.\ref{ViToutputFearture}). In ViT-B, as an alternative to raw image patches, the input sequence can be formed from feature maps of a CNN \cite{He_2016_CVPR}, the patch embedding projection $\mathsf{E}\in {{\mathbb{R}}^{({{m}^{2}}\cdot C)\times D}}$ is applied to patches extracted from a CNN feature map. The linear projection $\mathsf{E}$ maps $\mathsf{x}$ to $D$ dimensions. We refer to the output $\mathsf{y}$ of this projection as the patch embeddings:
\begin{equation}
\left[ {{\mathsf{y}}_{class}};{{\mathsf{y}}_{1}}\mathsf{;}{{\mathsf{y}}_{2}};\cdots ;{{\mathsf{y}}_{N}} \right]=\left[ {{\mathsf{x}}_{class}};\mathsf{x}_{m}^{1}\mathsf{E;x}_{m}^{2}\mathsf{E;}\cdots \mathsf{;x}_{m}^{N}\mathsf{E} \right]
  \label{beginprojection}
\end{equation}
Feed the sequence of patch embeddings to a standard $\mathsf{ViT}$ encoder: 
\begin{equation}
\left[ {{\mathsf{z}}_{class}};{{\mathsf{z}}_{1}};{{\mathsf{z}}_{2}};\cdots {{\mathsf{z}}_{N}} \right]=\underset{\theta }{\mathop{\mathsf{ViT}}}\,\left( \left[ {{\mathsf{y}}_{class}};{{\mathsf{y}}_{1}}\mathsf{;}{{\mathsf{y}}_{2}};\cdots ;{{\mathsf{y}}_{N}} \right] \right)
  \label{ViToutputFearture}
\end{equation}
where ${{\mathsf{z}}_{class}}$ is the image representation, which is input to the classifier to calculate the classification loss. $\theta $ denotes the trainable parameters of  $\mathsf{ViT}$. ${{\mathsf{z}}_{i}}=\left( {{\mathsf{z}}_{1}},\cdots ,{{\mathsf{z}}_{N}} \right)$ is the representation of the image patch $\mathsf{x}_{m}^{i}=\left( \mathsf{x}_{m}^{1},\cdots ,\mathsf{x}_{m}^{N} \right)$. We intercept half of ${{\mathsf{z}}_{i}}$, named $\frac{1}{2}{{\mathsf{z}}_{i}}$. Then use a trainable lightweight $\mathsf{MLP}$ block (shown in Figure 1) to project $\frac{1}{2}{{\mathsf{z}}_{i}}=\left( \frac{1}{2}{{\mathsf{z}}_{1}},\cdots ,\frac{1}{2}{{\mathsf{z}}_{N}} \right)$ as the $d$-dimensional positional feature vector:
\begin{equation}
\left[ {{\mathsf{p}}_{1}};{{\mathsf{p}}_{2}};\cdots ;{{\mathsf{p}}_{N}} \right]=\mathsf{MLP}\left( \left[ \frac{1}{2}{{\mathsf{z}}_{1}};\frac{1}{2}{{\mathsf{z}}_{2}};\cdots ;\frac{1}{2}{{\mathsf{z}}_{N}} \right] \right)
  \label{PositionOutputFeature}
\end{equation}
The probability that ${{\mathsf{p}}_{i}}$ belongs to ${{{y}}_{i}}$ by SoftMax is:
\begin{equation}
{{P}_{{{\mathsf{p}}_{i}},{{y}_{i}}}}=\frac{\exp (w_{{{y}_{i}}}^{\mathsf{T}}{{\mathsf{p}}_{i}})}{\sum\limits_{j=1}^{c}{\exp (w_{j}^{\mathsf{T}}{{\mathsf{p}}_{i}})}}
  \label{SoftMaxprobability}
\end{equation}
where $[{{\boldsymbol{w}}_{1}},\cdots ,{{\boldsymbol{w}}_{c}}]\in {{R}^{d\times c}}$ is the weights in the positional classiﬁer, $c$ denotes the number of classes (In ViT-B, $c=196$), and $d$ denotes the dimension of the feature. Each $\mathsf{x}_{m}^{i}$ corresponds to an absolute positional label ${{{y}}_{i}}=\left( {{{y}}_{1}},\cdots ,{{{y}}_{N}} \right)$, as shown in Figure \ref{absolute position-flabel} (${{{y}}_{1}}=\left( 0 \right),\cdots ,{{{y}}_{9}}=\left( 8 \right)$). Absolute positional label trains the model by minimizing cross-entropy:
\begin{equation}
{{L}_{p}}=-\frac{1}{n}\sum\limits_{i=1}^{n}{\log \left( {{P}_{{{\mathsf{p}}_{i}},{{y}_{i}}}} \right)}
\end{equation}

In the conventional classification task, we adopt the joint supervision of softmax loss and position loss to train the ViTs for discriminative feature learning. The formulation is given in Eq.\ref{AllLoss}.
\begin{equation}
L={{L}_{s}}+\lambda {{L}_{p}} 
 \label{AllLoss}
\end{equation}
where the scalar $\lambda$ is used for balancing the two loss functions.

\subsection{Relative Positional Label}

Besides the absolute position of each input image patch, we also consider the relative positional relationship between patches. Relative position methods encode the relative distance between input patches and learn the pairwise relationships of patches. We encode the relative position between the input image patches $\mathsf{x}_{m}^{i}$ and $\mathsf{x}_{m}^{j}$ into a vector $\mathsf{p}_{ij}^{r}$. To establish the pairwise positional relationship between the output image patch embeddings, we intercept half of ${{\mathsf{z}}_{i}}$, and combine them in pairs, as shown in Figure 2:
\begin{equation}
{{\mathsf{z}}_{ij}}=concat\left( \frac{1}{2}{{\mathsf{z}}_{i}},\frac{1}{2}{{\mathsf{z}}_{j}} \right)
  \label{eq:5}
\end{equation}
where the \emph{concat} operation concatenates the two embeddings. Same as the absolute positional label, feed ${{\mathsf{z}}_{ij}}$ to a lightweight MLP block:
\begin{equation}
\mathsf{p}_{ij}^{r}=\mathsf{MLP}\left( {{\mathsf{z}}_{ij}} \right)
  \label{eq:6}
\end{equation}
where $\mathsf{p}_{ij}^{r}$ denotes the relative positional feature vector between $\mathsf{x}_{m}^{i}$ and $\mathsf{x}_{m}^{j}$ output by the model. We use the method in Swin-ViT \cite{Liu_2021_ICCV} to convert 2D absolute positional labels to 1D relative position labels ${{y}_{ij}}$. Similar to Eq.4 and 5, the optimization goal of the relative positional label is:
\begin{equation}
L_{p}^{r}=-\frac{1}{{{n}^{2}}}\sum\limits_{i=1}^{n}{\sum\limits_{j=1}^{n}{\log \left( {{P}_{\mathsf{p}_{ij}^{r},{{y}_{ij}}}} \right)}}
  \label{eq:7}
\end{equation}

In the conventional classification task, the relative position loss is trained in the same way as Eq.\ref{AllLoss}.

\subsection{Coupling Positional Label and MAE}

\begin{figure}[h]
  \centering
   \includegraphics[width=1.0\linewidth]{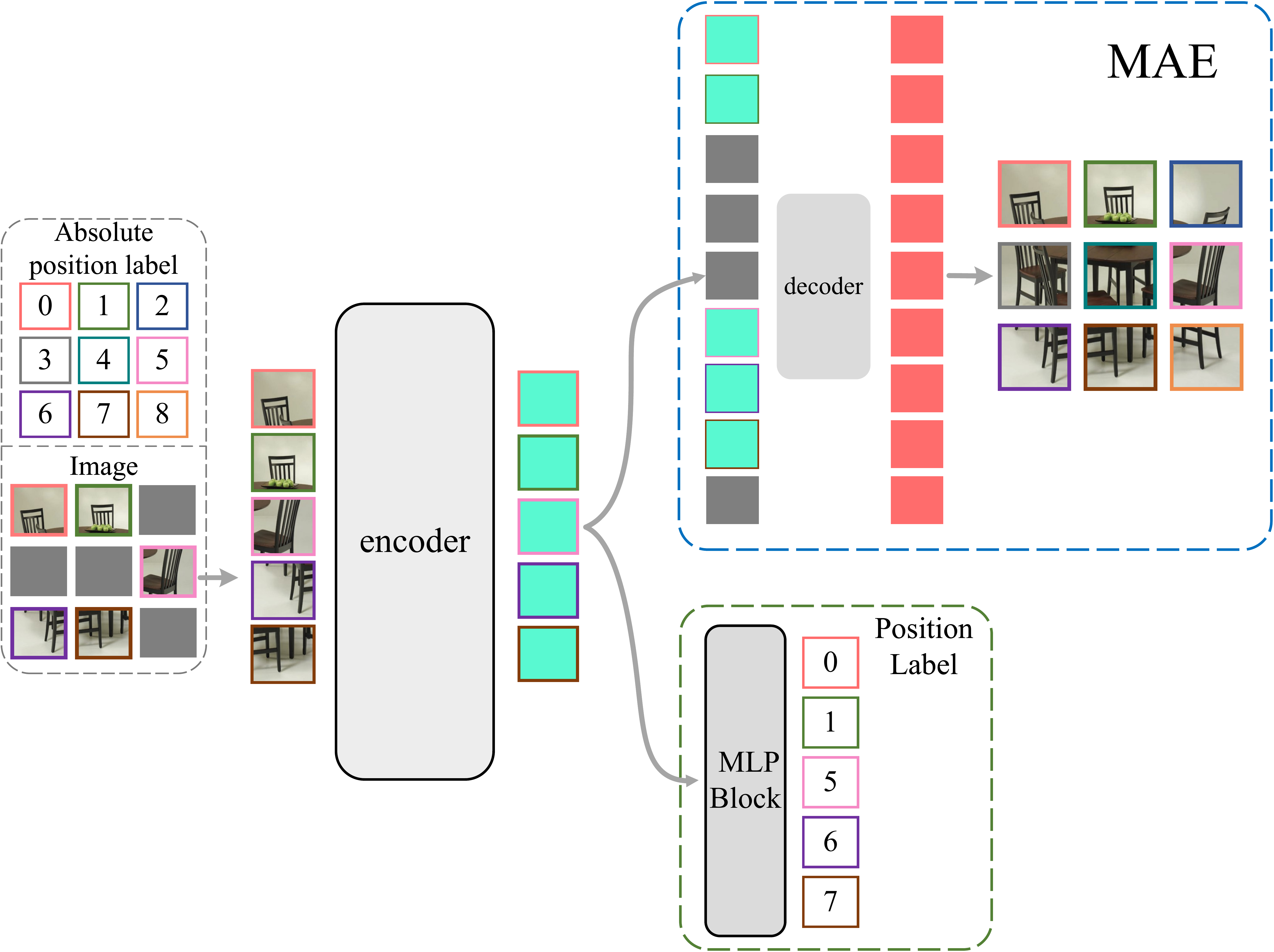}

   \caption{Our absolute positional label is combined with MAE. The output of the encoder is not only input to the decoder for MAE reconstruction of the image, but also to our positional label self-supervised MLP to output the absolute position of visible patches. This illustration was inspired by MAE.}
   \label{CombineMAE-flabel}
\end{figure}

The above two positional labels can not only be used as auxiliary supervision signals for classification tasks, but also can be used for self-supervised training of ViT, which can be combined with various current ViT self-supervised methods to obtain more powerful self-supervised signals, e.g., MAE \cite{DBLP:journals/corr/abs-2111-06377}. 

MAE masks a large random subset of image patches (e.g., 75\%). A small subset of visible patches is then fed into the encoder. Mask tokens are introduced after the encoder, and the full set of encoded patches and mask tokens is processed by a small decoder that reconstructs the original image in pixels. This section illustrates our positional label combined with MAE, as shown in Figure \ref{CombineMAE-flabel}. The output of the encoder is not only input to the decoder for MAE reconstruction of the image, but also to our positional label self-supervised MLP to output the absolute position of visible patches. Different from the original MAE, we do not add the encoder's positional encoding during pre-training, and only use our positional labels for self-supervised training to fuse the positional information. Add the encoder's positional encoding and use the proposed positional label, when fine-tuning the model.

\subsection{Coupling Positional Label and Hierarchical ViT}

Hierarchical ViT (e.g., Swin-ViT) to produce a hierarchical representation, the number of tokens is reduced by patch merging layers as the network gets deeper. The "Stage 3" and "Stage 4" of the original Swin-T, with output resolutions of $\frac{H}{16}\times \frac{W}{16}$ and $\frac{H}{32}\times \frac{W}{32}$, respectively, as shown in Figure \ref{modified SwinT-flabel}(a). Taking the input image resolution of $224\times 224$ as an example, the sequence length of the encoded patches output by the original Swin-T is $\frac{224}{32}\times \frac{224}{32}=49$, which means that the number of classes of the positional label classification task $c=49$, much smaller than the number of classes of ViT-B ($c=196$), which we found experimentally that this affects the performance of positional label. In order to adapt to the positional label, we slightly modified the architecture of Swin-ViT and removed the last patch merging layer, so that the output resolution of Swin-T's "Stage 4" is $\frac{H}{16}\times \frac{W}{16}$, keeping the number of classes of the positional label classification task $c=196$, as shown in Figure \ref{modified SwinT-flabel}(b). We found that this will not affect the performance of Swin-ViT, but it will increase the amount of calculation, which is also the limitation of the positional label in hierarchical ViT, and we will explore the solution to this problem in the next work. To efficiently incorporate positional labels, we use conditional positional encoding \cite{DBLP:journals/corr/abs-2102-10882} instead of the original relative positional bias. In the following experiments, we use the modified architecture for hierarchical ViT.

\section{Experiments}

We conduct experiments on ImageNet-1K \cite{5206848} image classification, Caltech-256 \cite{griffin2007caltech} and Mini-ImageNet \cite{krizhevsky2012imagenet} small datasets image classification. In the following, we first conduct ablation experiments to demonstrate that positional encoding can be combined with the proposed positional labels, and then combine the proposed positional labels with the latest ViT architecture to demonstrate the effectiveness of the positional labels. Finally, we combine the proposed positional labels with MAE to demonstrate the potential of positional labels in self-supervised ViT training.

\subsection{Experiment Settings}
\noindent \textbf{Dataset}. For image classification, we benchmark the proposed positional label on the ImageNet-1K, which contains 1.28M training images and 50K validation images from 1,000 classes. To explore the performance of positional label on small datasets, we also conducted experiments on Caltech-256 and Mini-ImageNet. Caltech-256 has 257 classes with more than 80 images in each class. Mini-ImageNet contains a total of 60,000 images from 100 classes.

\noindent \textbf{Implementation details}. This setting mostly follows \cite{Liu_2021_ICCV}. We use the PyTorch toolbox \cite{paszke2019pytorch} to implement all our experiments. We employ an AdamW \cite{kingma2014adam} optimizer for 300 epochs using a cosine decay learning rate scheduler and 20 epochs of linear warm-up. A batch size of 256, an initial learning rate of 0.001, and a weight decay of 0.05 are used. ViT-B/16 uses an image size 384 and others use 224. We include most of the augmentation and regularization strategies of \cite{Liu_2021_ICCV} in training.

\subsection{Ablation Studies}
\begin{table}[h]
   \centering
   \caption{Top-1 and Top-5 accuracies (\%) for ViT-B model, on the ImageNet dataset. PE: positional encoding term of ViT; APL: the proposed absolute positional label; RPL: the proposed relative positional label.}
   \begin{tabular}{l|c|c}
      \hline % \toprule
      Method                   & Top-1 acc.      & Top-5 acc.      \\
      \hline % \midrule
      ViT-B + PE (Baseline)      & 77.91     &92.48      \\
      ViT-B            & 76.34    &90.83      \\
      ViT-B + APL   &   78.19  &92.87\\
      ViT-B + RPL   &   78.00  &92.40\\
      ViT-B + PE + RPL   &   79.07  &93.67\\
     ViT-B + PE + APL  & \textbf{79.11} & \textbf{93.73}\\
      \hline % \bottomrule
   \end{tabular} \\
   \label{ViT-B-YesOrNoPositionCoding}
\end{table}

\begin{table}[h]
   \centering
   \caption{Top-1 and Top-5 accuracies (\%) for Swin-T model, on the ImageNet dataset.}
   \begin{tabular}{l|c|c}
      \hline % \toprule
      Method                   & Top-1 acc.      & Top-5 acc.      \\
      \hline % \midrule
      Swin-T + PE (Baseline)      & 81.32     &95.64      \\
      Swin-T            & 80.14    &94.93      \\
      Swin-T + APL   &   80.89  &95.17\\
      Swin-T + RPL   &   81.15  &95.30\\
      Swin-T + PE + RPL   &   \textbf{81.93}  &95.67\\
      Swin-T + PE + APL  & 81.51 & \textbf{95.83}\\
      \hline % \bottomrule
   \end{tabular} \\
   \label{Swin-T-YesOrNoPositionCoding}
\end{table}

\noindent \textbf{Positional encoding.} Incorporating explicit representations of positional information is known to be especially important for Transformers, since the model is otherwise entirely invariant to sequence ordering, which is undesirable for modeling structured data. Experiments show that our proposed positional label can completely replace positional encoding to add positional information to the ViT model. Surprisingly, we found that the proposed positional label can be used in combination with conventional positional encoding, the model does not learn an identity and can achieve the best performance gain. We experiment on ViT-B and Swin-T, as shown in Table \ref{ViT-B-YesOrNoPositionCoding} and Table \ref{Swin-T-YesOrNoPositionCoding}.

\begin{figure}[h]
  \centering
   \includegraphics[width=0.9\linewidth]{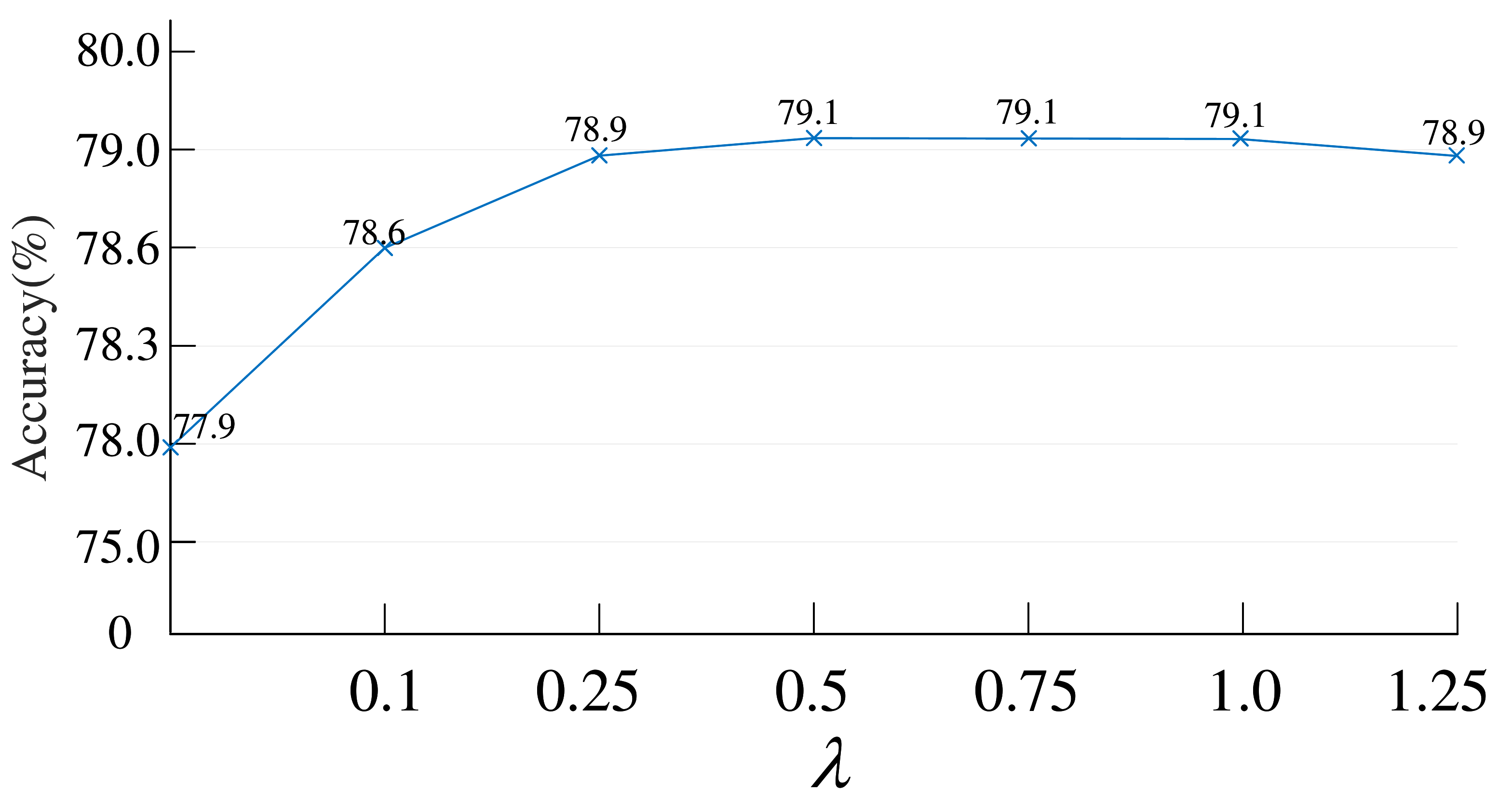}
   \caption{Illustration of position loss with different $\lambda$, using the ViT-B architecture.}
   \label{Hyperparameter-flabel}
\end{figure}

\noindent \textbf{Hyperparameter $\lambda$.} The hyperparameter $\lambda$ balances the classification loss and position loss. It is essential to our model. So we conduct an experiment to investigate the sensitiveness of this parameter. In our experiments, we vary $\lambda $ from 0 to 1.25 to learn different models. The Top-1 accuracies of these models on the ImageNet-1K dataset are shown in Figure \ref{Hyperparameter-flabel}. Properly choosing the value of $\lambda $ can improve the accuracy of the models. We also observe that the performance of the models remains largely stable across a wide range of $\lambda $. All experiments in this paper set $\lambda =0.5$.

\begin{figure}[h]
  \centering
   \includegraphics[width=1.0\linewidth]{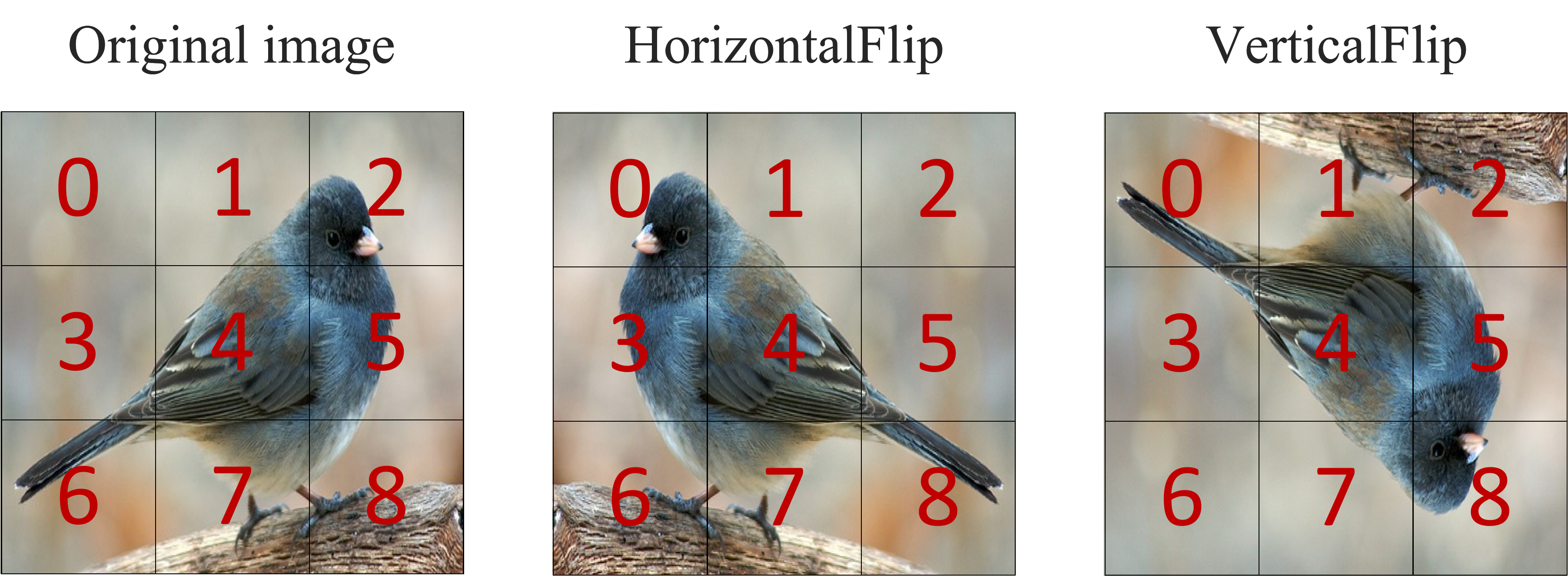}
   \caption{Data augmentation. Our positional label works with different image augmentations.}
   \label{ImageAugmentation-flabel}
\end{figure}

\noindent \textbf{Data augmentation.} Because the data augmentation affects our positional labels, as shown in Figure \ref{ImageAugmentation-flabel}, the image uses different data augmentations, and the bird head corresponds to different positional labels. Left: positional label is 2. Middle: positional label is 0. Right: positional label is 8. Table \ref{ViT-B-ImageAugmentation} studies the inﬂuence of data augmentation on our positional label. Using random horizontal ﬂipping for our positional label improves the performance, but combining random horizontal ﬂipping and random vertical ﬂipping has little effect on the results, suggesting that positional label does not require large augmentation to regularize training.

\begin{table}[h]
   \centering
   \caption{Data augmentation. Our positional label works with minimal augmentation, using the ViT-B architecture. crop: random resized crop; RHF: random horizontal ﬂipping; RVF: random vertical ﬂipping.}
   \begin{tabular}{l|c|c}
      \hline % \toprule
      Method                   & Top-1 acc.      & Top-5 acc.      \\
      \hline % \midrule
      crop + RHF (Baseline)      & 77.91     &92.48      \\
      crop + APL   &   78.54  & 93.13 \\
      crop + RPL   &   78.26  & 92.95 \\
      crop + RHF + APL   &   \textbf{79.11}  & \textbf{93.73} \\
      crop + RHF + RPL   &   79.07  &93.67\\
      crop + RHF + RVF + APL   &   79.09  &93.69\\
      crop + RHF + RVF + RPL   &   79.10  &92.87\\
      \hline % \bottomrule
   \end{tabular} \\
   \label{ViT-B-ImageAugmentation}
\end{table}

\subsection{Image Classiﬁcation on the ImageNet-1K}

\begin{table}[h]
   \centering
   \caption{Top-1 and Top-5 accuracies (\%) of different ViT backbones, on the ImageNet dataset. Each ViT variant follows the approach of adding positional encoding in its paper.}
   \begin{tabular}{l|c|c}
      \hline % \toprule
      Method                   & Top-1 acc.      & Top-5 acc.      \\
      \hline % \midrule
      ViT-B \cite{DBLP:journals/corr/abs-2010-11929}      & 77.91     &92.48     \\
      DeiT-B \cite{pmlr-v139-touvron21a}    & 81.87     &93.92     \\
      Swin-B \cite{Liu_2021_ICCV}    & 83.35     &\textbf{96.38}     \\
      NesT-B \cite{zhang2022nested}   & \textbf{83.67}     &96.16     \\
      \hline % \midrule
      ViT-B + APL      & 79.11     &93.73     \\
      DeiT-B + APL    & 82.49     &94.34     \\
      Swin-B + APL    & 83.76     &96.83     \\
      NesT-B + APL    & \textbf{84.13}     &\textbf{96.85}     \\
      \hline % \midrule
      ViT-B + RPL      & 79.07     &93.67     \\
      DeiT-B + RPL    & 82.85     &94.21     \\
      Swin-B + RPL    & \textbf{84.09}     &96.77     \\
      NesT-B + RPL    & 83.93     &\textbf{96.81}     \\
      \hline % \bottomrule
   \end{tabular} \\
   \label{ImageNet-Top1AndTop5}
\end{table}

We combine absolute and relative positional labels with various ViT variants, each ViT variant follows the approach of adding positional encoding in its paper. The experimental results show that the proposed positional label can significantly improve the performance of ViT, as shown in Table \ref{ImageNet-Top1AndTop5}: +1.20\% for ViT-B with the absolute positional label (79.11\%) over ViT-B (77.91\%), and +0.74\% for Swin-B with the relative positional label (84.09\%) over Swin-B (83.35\%). The experimental results show that the performance improvement of positional labels for the full-attention transformer (e.g., ViT-B) is higher than that of the local-attention transformer (e.g., Swin-B), which also indicates that the local-attention transformer can process limited data more efficiently. 

\begin{figure}[h]
  \centering
   \includegraphics[width=0.9\linewidth]{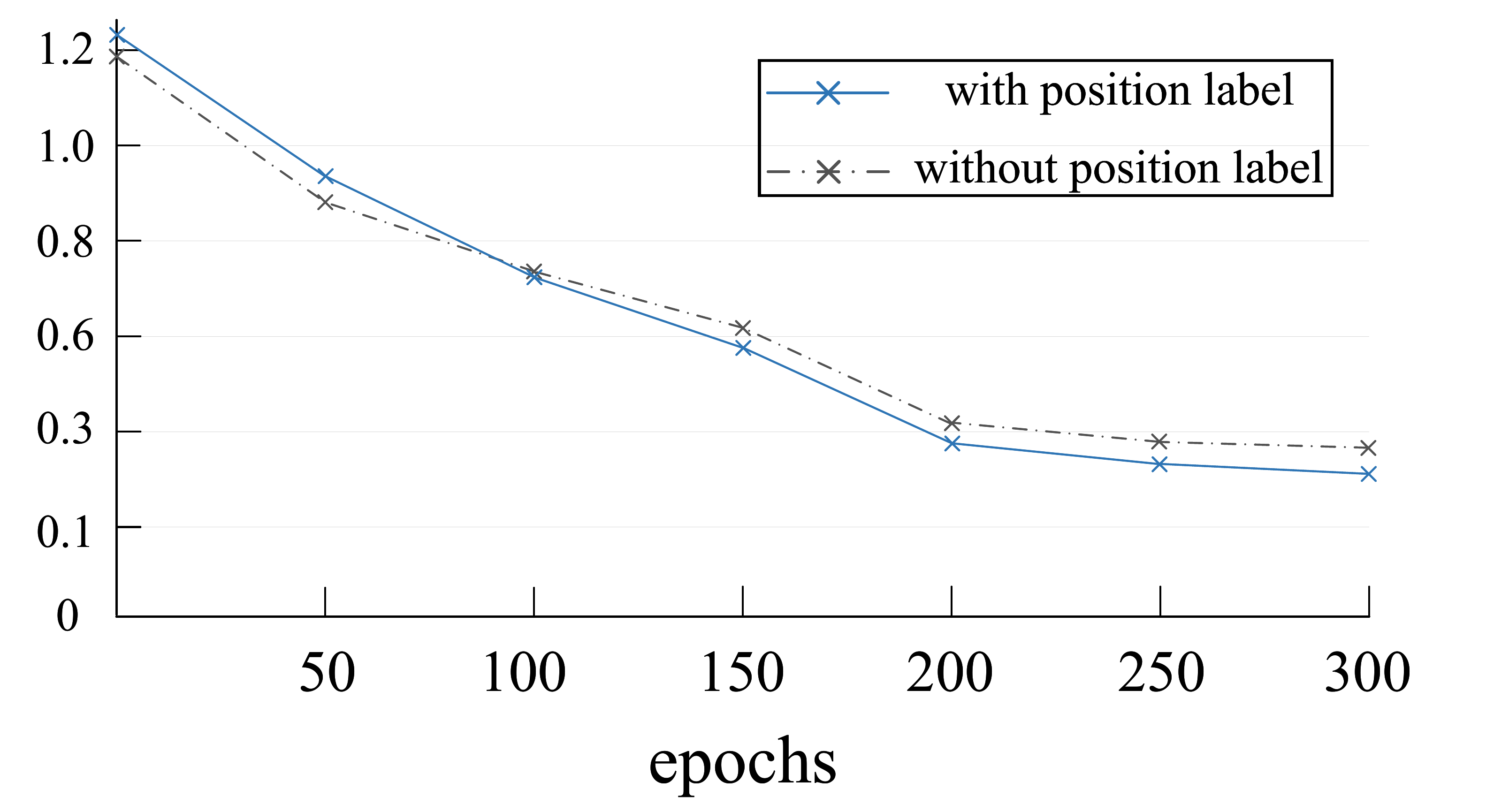}
   \caption{During training, the classification loss of ViT-B with the positional label and ViT-B without the positional label.}
   \label{TrainLoss-flabel}
\end{figure}

In the classification task, we use a classification loss function combined with our positional labels for joint training. We observe the classification loss changing curve of ViT and find an interesting phenomenon that the classification loss of ViT with positional labels is smaller (as shown in Figure \ref{TrainLoss-flabel}), which indicates that our positional labels can benefit the convergence of ViT.

\begin{table}[h]
   \centering
   \caption{Top-1 and Top-5 accuracies (\%) of various ViT variants, on the Mini-ImageNet dataset.}
   \begin{tabular}{l|c|c}
      \hline % \toprule
      Method                   & Top-1 acc.      & Top-5 acc.      \\
      \hline % \midrule
      ViT-B \cite{DBLP:journals/corr/abs-2010-11929}      & 58.28     &79.57     \\
      DeiT-B \cite{pmlr-v139-touvron21a}    & 63.67     &83.92     \\
      Swin-B \cite{Liu_2021_ICCV}    & 67.39     &\textbf{86.88}     \\
      NesT-B \cite{zhang2022nested}   & \textbf{67.43}     &86.75     \\
      \hline % \midrule
      ViT-B + APL      & 64.43     &83.73     \\
      DeiT-B + APL    & 66.49     &85.34     \\
      Swin-B + APL    & \textbf{68.91}     &87.83     \\
      NesT-B + APL    & 68.73     &\textbf{87.95}     \\
      \hline % \midrule
      ViT-B + RPL      & 63.97     &83.06     \\
      DeiT-B + RPL    & 66.85     &85.21     \\
      Swin-B + RPL    & 69.11     &88.02     \\
      NesT-B + RPL    & \textbf{69.56}     &\textbf{88.67}     \\
      \hline % \bottomrule
   \end{tabular} \\
   \label{Mini-ImageNet-Top1AndTop5}
\end{table}

\subsection{Image Classiﬁcation on Caltech-256 and Mini-ImageNet}

We show the performance of ViT on small datasets in Table \ref{Mini-ImageNet-Top1AndTop5} and Table \ref{Caltech-256-Top1AndTop5}. It is known that ViTs usually perform poorly on such tasks as they typically require large datasets to be trained on. The models that perform well on large-scale ImageNet do not necessary work perform on small-scale Mini-ImageNet and Caltech-256, e.g., ViT-B has top-1 accuracy of 58.28\% and Swin-B has top-1 accuracy of 67.39\% on the Mini-ImageNet, which suggests that ViTs are more challenging to train with less data. Our proposed positional label can accelerate the convergence of ViT and significantly improve the performance of ViT on small datasets like Mini-ImageNet and Caltech-256. On Mini-ImageNet, the top-1 accuracy of ViT-B and Swin-B are improved by 6.15\% and 1.72\%, respectively. On Caltech-256, the top-1 accuracy of ViT-B and Swin-B are improved by 3.16\% and 2.24\%, respectively.

\begin{table}[h]
   \centering
   \caption{Top-1 and Top-5 accuracies (\%) of various ViT variants, on Caltech-256 dataset.}
   \begin{tabular}{l|c|c}
      \hline % \toprule
      Method                   & Top-1 acc.      & Top-5 acc.      \\
      \hline % \midrule
      ViT-B \cite{DBLP:journals/corr/abs-2010-11929}      & 37.57     &56.83     \\
      DeiT-B \cite{pmlr-v139-touvron21a}    & 40.56     &61.24     \\
      Swin-B \cite{Liu_2021_ICCV}    & \textbf{46.67}     &\textbf{67.22}     \\
      NesT-B \cite{zhang2022nested}   & 45.54     &66.35     \\
      \hline % \midrule
      ViT-B + APL      & 40.73     &60.63     \\
      DeiT-B + APL    & 41.79     &62.86     \\
      Swin-B + APL    & 48.91     &68.33     \\
      NesT-B + APL    & \textbf{49.77}     &\textbf{68.84}     \\
      \hline % \midrule
      ViT-B + RPL      & 40.51     &61.07     \\
      DeiT-B + RPL    & 41.28     &61.56     \\
      Swin-B + RPL    & 48.80     & \textbf{68.09}     \\
      NesT-B + RPL    & \textbf{48.97}     &67.99     \\
      \hline % \bottomrule
   \end{tabular} \\
   \label{Caltech-256-Top1AndTop5}
\end{table}

\subsection{Coupling Positional Label and MAE}

\begin{table}[h]
   \centering
   \caption{Experimental results of positional label combined with MAE on the ImageNet-1K. Pre-trained for 400 epochs.}
   \begin{tabular}{l|c}
      \hline % \toprule
      Method                   & Top-1 acc.           \\
      \hline % \midrule
      ViT-B \cite{DBLP:journals/corr/abs-2010-11929}      & 77.91         \\
      ViT-B + MAE \cite{DBLP:journals/corr/abs-2111-06377}      & 79.54        \\
      ViT-B + MAE + APL   &   \textbf{80.40}   \\
      ViT-B + MAE + RPL   &   80.21   \\
      \hline % \bottomrule
   \end{tabular} \\
   \label{MAE-accuracy}
\end{table}

We do self-supervised pre-training on the ImageNet-1K training set. Then we do supervised training to evaluate the representations with end-to-end fine-tuning. We report top-1 validation accuracy. This experiment uses ViT-B as the backbone, and the experimental setting follows the default setting of MAE \cite{DBLP:journals/corr/abs-2111-06377}. Table \ref{MAE-accuracy} is a comparison between ViT-B trained from scratch \emph{vs.} ﬁne-tuned from MAE with our positional label.

\begin{figure}[h]
  \centering
   \includegraphics[width=0.9\linewidth]{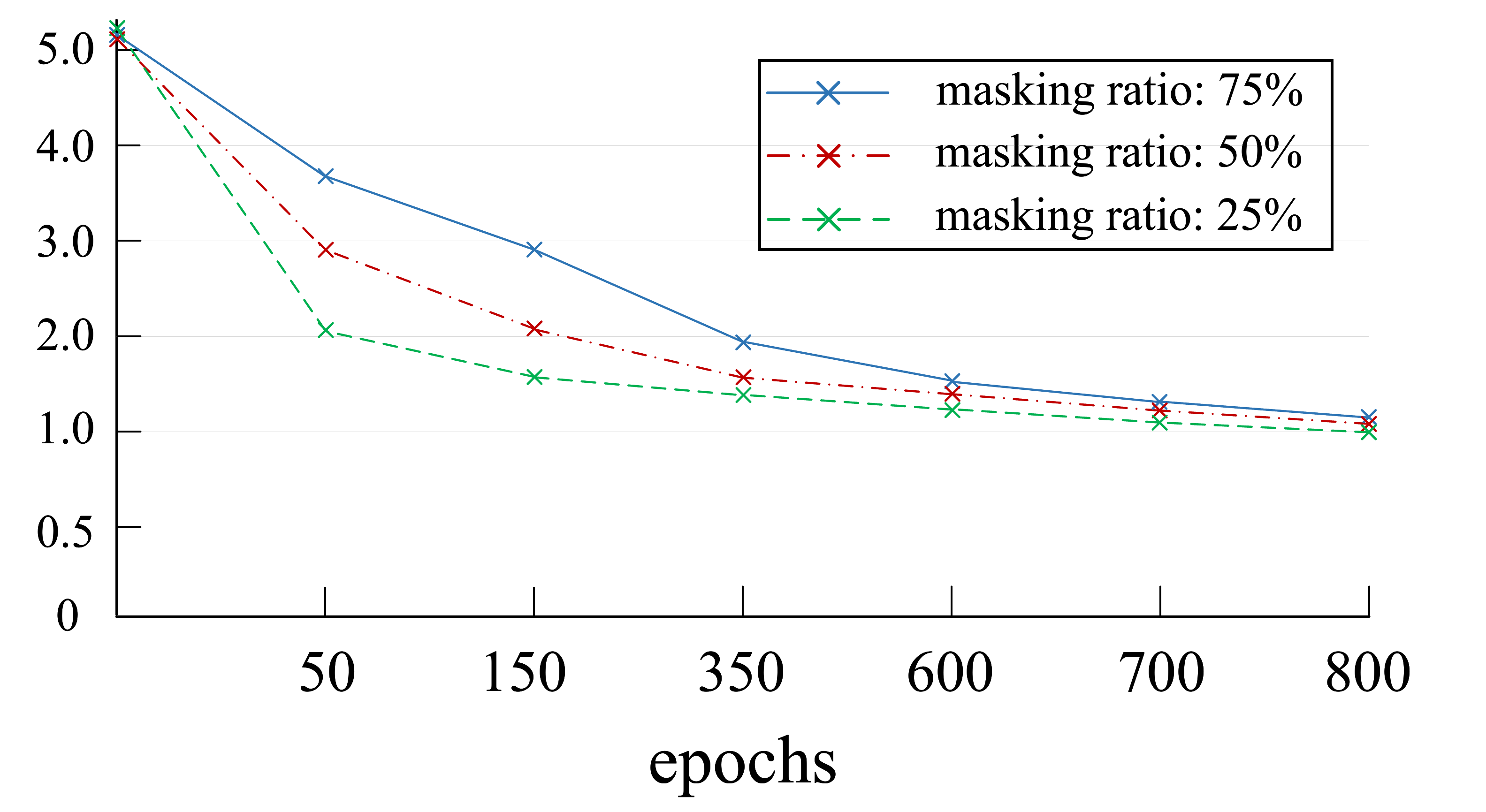}
   \caption{The position loss with different masking ratios.}
   \label{MaskRatio-flabel}
\end{figure}

MAE finds that masking a high proportion of the input image, e.g., 75\%, yields a nontrivial and meaningful self-supervisory task. Random sampling with a high masking ratio largely eliminates redundancy, thus creating a task that cannot be easily solved by extrapolation from visible neighboring patches. We find that this also applies to our proposed positional label, where the encoder uses a small subset of visible patches to reconstruct the overall positional information of the image, yielding a more powerful self-supervised signal. As shown in Figure \ref{MaskRatio-flabel}, the training difficulty of the model increases as the masking ratio increases.

Under this design, MAE and our positional label can achieve a win-win scenario: MAE reconstructs clear masked patch pixels that help the positional loss capture the overall positional information of the image, while the positional loss also incorporates the positional information into the MAE reconstruction, allowing the model to consider the overall structure of the image during the reconstruction process.

\section{Conclusions}

In this paper, we propose two positional labels for self-supervised training of ViT. The extensive experiments show that our method can be combined with various ViT variants, bringing significant improvements on classification tasks. Our methods could be easily plugged into self-attention layers. In addition, our positional labels can also be used for ViT fully self-supervised methods to provide a powerful self-supervised signal.

\bibliographystyle{named}
\bibliography{ijcai22}

\end{document}